\documentclass{clv3}
\usepackage[T2A,T1]{fontenc}
\usepackage[utf8]{inputenc}
\usepackage[russian,english]{babel}
\usepackage{slantsc}
\usepackage{natbib}
\usepackage{hyperref}
\usepackage{multirow}
\usepackage{xcolor}
\usepackage{booktabs}
\usepackage{colortbl} 
\usepackage{appendix}
\usepackage[position=b]{subcaption}
\definecolor{darkblue}{rgb}{0, 0, 0.5}
\hypersetup{colorlinks=true,citecolor=darkblue, linkcolor=darkblue, urlcolor=darkblue}
\newcommand{\todo}[1]{\textcolor{blue}{\textbf{TODO: #1}}}
\renewcommand{\todo}[1]{}
\usepackage{amsmath}
\usepackage{arydshln}

\DeclareMathOperator*{\argmax}{argmax} % thin space, limits underneath in displays

\dochead{}

\runningtitle{Semantic Drift in Multilingual Representations}

\runningauthor{Beinborn \& Choenni}

\begin{document}

\title{Semantic Drift in Multilingual Representations}

\author{Lisa Beinborn}
\affil{Vrije Universiteit Amsterdam, E-mail: l.m.beinborn@vu.nl}
\author{Rochelle Choenni}
\affil{Universiteit van Amsterdam, E-mail: rochelle.choenni@student.uva.nl}

\maketitle

\begin{abstract}
Multilingual representations have mostly been evaluated based on their performance on specific tasks. In this article, we look beyond engineering goals and analyze the relations between languages in computational representations. We introduce a methodology for comparing languages based on their organization of semantic concepts. We propose to conduct an adapted version of representational similarity analysis of a selected set of concepts in computational multilingual representations. Using this analysis method, we can reconstruct a phylogenetic tree that closely resembles those assumed by linguistic experts. These results indicate that multilingual distributional representations which are only trained on monolingual text and bilingual dictionaries preserve relations between languages without the need for any etymological information. In addition, we propose a measure to identify semantic drift between language families. We perform experiments on word-based and sentence-based multilingual models and provide both quantitative results and qualitative examples. Analyses of semantic drift in multilingual representations can serve two purposes: they can indicate unwanted characteristics of the computational models and they provide a quantitative means to study linguistic phenomena across languages.
\end{abstract}
\section{Introduction}
Aligning the meaning of multiple languages in a joint representation to overcome language barriers has challenged humankind for centuries. Multilingual analyses range from the first known parallel texts on Rosetta stone over centuries of lexicographic work on dictionaries to online collaborative resources like \textsc{Wiktionary} \cite{Meyer2012} and \textsc{Babelnet} \cite{Navigli2010}. These resources vary in their semantic representations, but they rely mostly on symbolic approaches such as glosses, relations, and examples. In the last decade, it has become a common standard in natural language processing to take a distributional perspective and represent words, phrases, and sentences as vectors in high-dimensional semantic space. These vectors are learned based on co-occurrence patterns in corpora with the objective that similar words should be represented by neighboring vectors. For example, we expect \textit{table} and \textit{desk} to appear close to each other in the vector space.  

Recently, approaches to unifying these monolingual semantic representations into a joint multilingual semantic space have become very successful \cite{Klementiev2012,Vulic2016,Conneau2018}. The goal is to assign similar vectors to words that are translations of each other without affecting the monolingual semantic relations between words. For example, \textit{table} should appear close to its Italian translation \textit{tavola} without losing the proximity to \textit{desk} which should in turn be close to the Italian \textit{scrittoio}. 

Cognitively inspired analyses have shown that the semantic organization of concepts varies between languages and that this variation correlates with cultural and geographic distances between language families \cite{Thompson2018quantifying,eger2016language}. We define this phenomenon as multilingual semantic drift and analyze to which extent it is captured in multilingual distributional representations. To this end, we propose a methodology for quantifying it which is based on the neuroscientific method of representational similarity analysis. Our approach uses a selected set of concepts and estimates how monolingual semantic similarity between concepts correlates across languages. We find that the results from our data-driven semantic method can be used to reconstruct language trees that are comparable to those informed by etymological research. We perform experiments on word-based and sentence-based multilingual models and provide both quantitative results and qualitative examples.

% Word-based multilingual models balance monolingual semantic similarity and crosslingual translation constraints and optimize them jointly whereas many sentence-based multilingual models are only optimized for translation. We examine whether these differences in the training objective have an influence on 

The article first introduces the most common architectures for multilingual distributional representations of words and sentences and then discusses approaches for quantifying the semantic structure that emerges in these models. These computational methods can be used to determine phylogenetic relations between languages. We elaborate on the data, the models, and the details of our analysis methods in an extensive methodology section. In a pilot experiment, we first evaluate the translation quality of the models for our datasets. The remaining sections discuss the results for the representational similarity analysis, the language clustering and the identification of semantic drift.  The code is available at \textnormal{\url{https://github.com/beinborn/SemanticDrift}}.

\section{Multilingual Distributional Representations}\label{s:relatedwork_models}
The large success of monolingual distributional representations of words gave rise to the development of representations for longer sequences such as phrases and sentences. Researchers soon moved from monolingual to multilingual space and developed methods to obtain comparable representations for multiple languages. In this section, we introduce the related work for creating multilingual representations for words and sentences, and discuss approaches for capturing semantic structure and phylogenetic relations. 
\subsection{Multilingual Representations for Words}
Approaches for constructing multilingual representations for words can be distinguished into two main classes: mapping models and joint models \cite{Ruder2017}.\footnote{\citet{Ruder2017} describe a third class of models which they call "Pseudo-multilingual corpora-based approaches". They then show that these models are mathematically similar to the mapping models.} The multilingual modeling techniques are very similar to those applied on learning shared representations for multiple modalities, e.g., vision and language \cite{beinborn2018multimodal,Baltrusaitis2019}. 

\paragraph{Mapping models} Mapping approaches are based on pre-trained monolingual representations of two languages (the source and the target language) and aim to project the representations from the semantic space of the source language to the target space. This approach is based on the idea that the intralingual semantic relations are similar across languages \cite{Mikolov2013b} and can be exploited to learn a linear projection from one language to the other. The linear projection is learned based on a bilingual seed dictionary that provides a link between the semantic spaces \cite{Vulic2016, Gouws2015mapping} or by aligning information from parallel corpora. In general, mapping models are directional and map representations from one language to the other. \citet{Faruqui2014} propose to instead map both representations into a joint space by applying canonical correlation analysis. During training, they enforce maximum correlation of representations for words that are known to be translations of each other.  

\paragraph{Joint models} For joint approaches, both representations are learned simultaneously by using parallel corpora for training. These models jointly optimize the objectives for monolingual and cross-lingual similarity. The monolingual objective is based on co-occurrence patterns observed in context and is similar to those that are commonly applied for training monolingual representations, e.g., the skip-gram objective in \textsc{word2vec} \cite{mikolov2013} or variants of it \cite{Luong2015}. The cross-lingual objective can be derived from word alignments \cite{Klementiev2012}, sentence alignments \cite{Gouws2015Bilbowa}, or document alignments \cite{Sogaard2016, Fung1998}. \\

Most of the models described above are inherently bilingual rather than multilingual. \citet{Duong2017} and \citet{Levy2017} show that learning representations for multiple languages simultaneously is beneficial because it facilitates transfer learning between closely related languages. We refer the interested reader to the detailed survey by \citet{Ruder2017} for further explanations on the mathematical foundations of cross-lingual representations.

The quality of a multilingual model is dependent on the quality of the cross-lingual signal. Several approaches aimed at enriching the signal by incorporating additional resources, such as visual cues  \cite{bergsma2011,Vulic2016multimodal} or syntactic information \cite{Duong2015pos}. Unfortunately, aligned multilingual corpora are usually scarce in low-resource languages. For covering a wider range of languages, self-supervised approaches which do not rely on predefined alignments have been developed. 

\paragraph{Self-supervised models}  \citet{Smith2017} and \citet{Hauer2017} derive the cross-lingual information by simply exploiting identical character strings from loanwords or cognates. As this only works for languages with the same alphabet, \citet{Artexe2017} go one step further and instantiate their model only with aligned digits. \citet{Conneau2018} and \citet{Zhang2017} do not use any parallel data and apply adversarial training to optimize the mapping between languages. Their generator tries to map the source words into the target space, while the discriminator attempts to distinguish between the target representations and the mapped source representations. As both the discriminator and the generator get better at their task, the mapped representations resemble more the target representations.  More recently, their approach has been transformed into a generative model using variational autoencoders \cite{Dou2018}.

In this work, we use the \textsc{Muse} model which has become popular due to its performance in self-supervised settings \cite{Conneau2018}. The model is based on monolingual representations (\textsc{fasttext}) which are trained on merged text data from \textsc{Wikipedia} and the \textsc{CommonCrawl} corpus and obtain good results for a wide range of languages \cite{fasttext}. While the \textsc{Wikipedia} data alone might contain a small domain bias because the articles cover varying ranges of topics across languages, the immense size of the \textsc{CommonCrawl} corpus provides a good approximation of actual written language use. It contains 24 terabytes of raw text data crawled from the web summing up to several billions of tokens for each language \cite{grave2018}. In order to ensure high quality for our experiments, we rely on multilingual representations obtained in a supervised fashion using a ground-truth bilingual dictionary.\footnote{See \url{https://github.com/facebookresearch/muse/} for details.} The entries of the dictionary serve as anchor points to learn a mapping from the source to the target space which is optimized by Procrustes alignment \cite{Procrustes}.

\subsection{Multilingual Representations for Sentences}\label{s:sentencerep}
The need for developing multilingual representations for sentences is most prevalent in the field of machine translation. Already in the 1950s, the idea of an interlingua  that could serve as a bridge between multiple languages emerged \cite{Gode1951}. The idea was further pursued by searching for a formalism that should represent the semantic content of a sentence independent of the language in which it is realized \cite{Richens1958}. Similar ideas have driven the development of logic formalisms such as Montague grammars \cite{Montague1970}. With the incredible success of powerful neural networks, it has currently become widely accepted that the most suitable form for such inter-lingual or multilingual representations are high-dimensional vectors. 

While discussing the wide range of machine translation literature is beyond the scope of this article, we briefly describe two main state-of-the-art models: encoder-decoder architectures and the transformer architecture. 

\paragraph{Encoder-Decoder} Machine translation is commonly interpreted as a sequence-to-sequence learning problem.  \citet{Sutskever2014} paved the way for fast developments on so-called encoder-decoder architectures. The encoder reads the input and learns to transform it into an intermediate representation that is then fed to the decoder to generate the translation of the sentence in a target language. Both the encoder and the decoder can be realized as different types of recurrent neural networks and can be combined with different techniques of attention \cite{Bahdanau2014}. Recently, bi-directional long short-term memory networks (BiLSTMs) have proven to be a good choice for modelling language \cite{peters2018}. \citet{Schwenk2018acl} show that using a joint BiLSTM encoder for all input languages combined with max-pooling over the last layer yields more robust sentence representations. After training, the decoder that is responsible for the translation generation is discarded and the output of the trained encoder is used as universal sentence representation.  These sentence representations can be interpreted as "sort of a continuous space interlingua" \cite[p.158]{Schwenk2017}. We use a pre-trained version of this model which is called \textsc{Laser} \cite{Artetxe2018}. 

\paragraph{Transformer} More recently, \citet{Vaswani2017} introduced the transformer model as a more sophisticated architecture for sequence to sequence transduction. Its underlying architecture follows the encoder-decoder paradigm, but no recurrent connections between tokens are used which reduces the training time for the model. In order to capture relations between tokens, a complex attention mechanism called multi-headed self-attention is applied and combined with positional encoding for signaling the order of tokens. Due to its success, variants of the transformer model for machine translation are currently being developed in a very fast pace. In the past, language modelling has commonly been interpreted as a left-to-right task, similar to incremental human language processing \cite{Rosenfeld2000}. As a consequence, the self-attention layer could only attend to previous tokens. \citet{Bert} argue that this approach unnecessarily limits the expressivity of the sentence representation. They propose to change the training objective from predicting the next word to predicting a randomly masked token in the sentence by considering both the left and right context. This task is also known as the \textit{cloze} task \cite{Taylor1953}. \citet{Bert} use this training objective to train a multi-layer bidirectional transformer (called \textsc{Bert}) and find that it strongly outperforms the previous state of the art on the \textsc{GLUE} evaluation corpus \cite{Glue}. By now, they have also released a multilingual version of \textsc{Bert} for 104 languages.\footnote{\url{https://github.com/google-research/bert/}}  

\textsc{Bert} and \textsc{Laser} obtain comparable results on the cross-lingual entailment dataset \cite{Conneau2017xnli}. 
For this article, we decided to use \textsc{Laser} because the model already outputs sentence representations that have a uniform dimensionality independent of the length of the sentence. This makes it possible to avoid additional experimental parameters for scaling the dimensionality of the sentence representations. The model has been trained by combining multiple multilingual parallel corpora from the \textsc{Opus} website \cite{tiedemann2012} accumulating to a total of 223 million parallel sentences \cite{Artetxe2018}.\footnote{The authors combined multilingual corpora with the aim to balance formal and informal language and long and short sentences. See Appendix A for details on the training data.} Note that the sentence-based model is optimized for translation whereas the word-based model aims at optimizing both monolingual semantic similarity and cross-lingual translation constraints. These different training objectives might have an influence on the model's ability to capture semantic differences between languages.

\subsection{Semantic Structure in Multilingual Representations}
Multilingual representations are commonly evaluated based on their performance on downstream tasks such as bilingual lexicon induction \cite{Vulic2016} and machine translation \cite{Zou2013}. More indirectly, multilingual representations are used for cross-lingual transfer in tasks such as information retrieval, or document classification \cite{Klementiev2012}. From a semantic perspective, multilingual representations are evaluated by comparing distances in the vector space with cross-lingual semantic similarity judgments by humans \cite{Semeval2017task}. Sentence representations are often tested by their ability to distinguish entailment relations between sentences \cite{Conneau2017xnli}. Most of these evaluations are simply multilingual extensions of monolingual evaluation tasks. These tasks ignore an important aspect of multilingual representations, namely the relations between languages.  

\paragraph{Phylogenetic relations}
Languages are fluid cultural constructs for communication which undergo continuous finegrained structural transformation due to emergent changes in their usage. A large body of work in historical linguistic aims to quantify how languages evolve over time and how different languages are related to each other. For example, Italian and Spanish both evolved from Latin and are thus more similar to each other than to Eastern European languages like Polish. One way of visualizing the typological relations between languages are phylogenetic trees. As a complement to historical research, computational analysis methods aim to automatically reconstruct phylogenetic relations between languages based on measurable linguistic patterns. We shortly introduce three main approaches for measuring language similarity: 
\begin{enumerate}
    \item \textit{Lexical overlap:} Earlier work on reconstructing phylogenetic relations mostly relies on determining lexical overlap between languages based on manually assembled lists of cognates \cite{Nouri2016} which is a cumbersome and subjective procedure \cite{Geisler2010}. Several methods for automatically extracting cognates exist (e.g., \citet{Serva2008}), but these approaches rely on the surface structure of a word. \citet{beinborn2013cognate} use character-based machine translation to identify cognates based on regular production processes, but their method still cannot capture the cognateness between the English \textit{father} and the Italian \textit{padre}, for example. For the methodology that we propose here, we abstract from the surface appearance of the word and focus on its semantic properties. As a consequence, we do not need to transliterate words from languages with different alphabets. 
    \item \textit{Structural similarity:} The similarity between languages is often measured by the similarity of their structural properties \cite{Cysouw2013}. The World Atlas of Language Structures (\textsc{wals}) lists a large inventory of structural properties of languages including phonological, grammatical, and lexical features.\footnote{\url{https://wals.info/}} \citet{Rabinovich2017} analyze translation choices and find that the syntactic structure of the source language is reflected in English translations. Recently, \citet{bjerva2019language} build on their work and analyzed the structural similarity between languages using phrase structure trees and dependency relations. Both approaches are able to reconstruct a phylogenetic tree solely based on syntactic features of the translation. We apply the same evaluation method for estimating the quality of the generated tree in section \ref{s:quantitative}, but we estimate the similarity of languages based on their semantic organization.
    \item \textit{Semantic organization:} Recent works indicate that semantic similarity between languages can also serve as a proxy for determining language families. \citet{eger2016language} find that semantic similarity between languages correlates with the geographic distance between countries in which the languages are spoken. In a similar vein, \citet{Thompson2018quantifying} find that semantic similarity between languages is proportional to their cultural distance. In these works, semantic structure is approximated by graph-based networks which indicate the associative strength between words. The challenge lies in aligning these structures across languages and accounting for complex semantic phenomena such as polysemy and context sensitivity \cite{youn2016universal}.
\end{enumerate}

\noindent Distributional representations ground computational models of language on the context patterns observed in corpora and enable us to quantify the semantic organization of concepts based on their distance in high-dimensional semantic space. These quantitative accounts of semantic structure facilitate the analysis of semantic phenomena such as monolingual semantic drift over time and multilingual semantic drift over language families.  

\paragraph{Semantic drift}  Semantic drift is mostly known from diachronic studies where it indicates the change of meaning over time \cite{Li2019,Frermann2016,Hamilton2016diachronic}.\footnote{The distinction between the terms \textit{semantic shift}, \textit{semantic change}, and \textit{semantic drift} is blurry in the literature. We are using \textit{semantic drift} here similar to \citet{Li2019} because \textit{shift} and \textit{change} tend to be used for characterizing more conscious processes.} Popular examples are the meaning drift of \textit{gay} from \textit{cheerful} to \textit{homosexual} over the years or the transformation of cognates into false friends as in \textit{gift} which today means \textit{poison} in German (but originally referred to \textit{something given}). Recently, a wide range of distributional approaches have been developed for measuring diachronic semantic change (see surveys by \citet{Tahmasebi2018survey} and \citet{Kutuzov2018survey} for an overview). 

\paragraph{Multilingual semantic drift} Semantic drift can also be observed across languages because even an exact translation of a word or a phrase does not share all semantic associations. For example, \textit{pupil} could be translated to Spanish as \textit{pupila}, but the Spanish phrase would only be associated with the eye and not with school children. These differences in the semantic scope of a word can lead to important differences in translation. \citet{Conneau2017xnli} observe that the English term \textit{upright} had been translated to Chinese as \textit{sitting upright}. As a consequence, the original sentence entailed \textit{standing} in their multilingual entailment corpus, but the translation violated this entailment relation. In this work, we analyze to which extent multilingual models preserve these semantic drifts. \citet{Faruqui2014} claim that multilingual projection can contribute to word sense disambiguation. For example, the polysemous English word \textit{table} is translated to \textit{tafel} in Dutch if it refers to a kitchen table, and to \textit{tabel} if it refers to a calculation matrix.
They provide a qualitative example for the word \textit{beautiful} to show that synonyms (\textit{pretty}, \textit{charming}) and antonyms (\textit{ugly}, \textit{awful}) are better separated in multilingual spaces.  \citet{Dinu2014} analyze zero-shot learning in multilingual and multimodal models and conversely find that fine-grained semantic properties tend to be washed out in joint semantic space. They describe the "hubness problem" as the phenomenon that a few words (the hubs) occur among the nearest neighbors for a large number of other words and show that this problem is more severe in mapped representational spaces.\footnote{\citet{Lazaridou2015} find that applying max-margin estimation instead of ridge regression for the mapping reduces the problem.}

\paragraph{Comparing semantic structure}
The approaches for comparing semantic structure over time and over languages are similar. Concepts are represented as vectors in high-dimensional semantic space and can be compared diachronically by calculating different vectors for each time epoch using historical corpora \cite{Hamilton2016diachronic,Rosenfeld2018} or multilingually by calculating different vectors for each language \cite{asgari2016comparing}. As the global position of a vector is often not comparable across corpora, the semantic drift is approximated by determining changes in the relative position of a concept within its local neighborhood \cite{Hamilton2016}. The resulting semantic networks can be compared by representing them as graph structures \cite{eger2016language,youn2016universal} or second-order similarity matrices \cite{Hamilton2016, Thompson2018quantifying}. The distance between these computational structures can then be used as an indicator for the amount of drift. An additional challenge for multilingual comparisons lies in determining the alignment of concepts across languages. Our computational approach is most similar to the method by \citet{Thompson2018quantifying}. They use a much larger set of stimuli (more than 1,000) for which gold translations are available and analyze their representations in monolingual embedding models. In our work, we focus on multilingual representations and analyze the cross-lingual similarity that emerges from the model. We extract translations in a data-driven way by taking the nearest semantic neighbor in semantic space instead of relying on translation resources. The details of our methodology are described in the following section.

% Recent works in natural language processing aim to quantify the semantic similarity between two languages by comparing their semantic organization of concepts. \citet{asgari2016comparing} calculate graph structures to represent the semantic organization of a language and use the distance between graphs as an indicator for the phylogenetic distance between languages. Their calculations are based on monolingual word representations and sentence alignments obtained from bible corpora. \citet{eger2016language} take a similar approach and use bilingual word representations. 

% Eger analyze bilingual word representations to examine whether the semantic similarity of the two languages can have an effect on the quality of the representations and obtain mixed results. 
 
 \section{Methodology}\label{s:methodology}
 In this section, we detail the methodology applied for our experiments. We provide information on the data, the multilingual models, and the methods used for comparing representational spaces. 
 
\subsection{Data}
We perform our word-based experiments with a set of stimuli which have been selected to be universal representatives of the most important semantic concepts. For the sentence-based experiments, we extract sentences from a parallel corpus. More information on the data and the languages can be found in the Appendix. As we are using very common and frequent data, it is likely that the stimuli have also occurred in the training data of the models. However, we are interested in examining the resulting representational relations between stimuli and the effect of the multilingual training regime of the models. 

\subsubsection{\textsc{Swadesh} Words}
The American linguist Morris Swadesh composed several lists of so-called language universals: semantic concepts that are represented in all languages \cite{Swadesh1955}. His lists have been revised multiple times and have also been subject to strong criticism \cite{Starostin2013,Geisler2010}. Nevertheless, they are still a popular tool in comparative linguistics and have been collected for a large range of languages and dialects. We are using the extended list of 205 different English words that is available on Wiktionary \cite{Swadeshlist}, see Appendix C.

\subsubsection{\textsc{Pereira} Words} \citet{Pereira2018} selected semantic concepts by performing spectral clustering on word representations obtained from \textsc{Glove} \cite{Glove}. They selected concepts by maximizing the variation on each dimension of the semantic space. After pre-processing, they manually selected 180 words (128 nouns, 22 verbs, 23 adjectives, 6 adverbs, 1 function word) claiming that their selection best covers the semantic space (see Appendix D). The concepts were originally selected to serve as stimuli in brain imaging experiments on semantic language processing in humans. The \textsc{Pereira} list overlaps with the \textsc{Swadesh} list for 20 words. We ignore the word \textit{argumentatively} because it is not in the vocabulary of the \textsc{Muse} model.  
\subsubsection{\textsc{Europarl} Sentences} \citet{Koehn2005} extracted the \textsc{Europarl} corpus from the proceedings of the European Parliament. It includes sentence-aligned versions in 21 European languages. As the corpus contains formal political language, short sentences are often captions or names of documents and long sentences tend to be overly complex. 
We thus decided to restrict our analysis to sentences of medium length ranging from 6 to 20 words. In order to better control for sentence length, we extract three sets of 200 random sentences each conforming to three length constraints. The set of \textit{short} sentences consist of 6--10 words, \textit{mid} sentences of 11--15 words and \textit{long} sentences of 16--20 words. We restrict the 21 languages to the 17 used in \citet{Rabinovich2017}. Whereas they use only English sentences (which are translations) and examine the syntactic structure of the sentence with respect to the language of the source sentence, we use translations into multiple languages and keep the set of sentences constant. 

\subsection{Multilingual Models}
We use two different freely-available pre-trained multilingual representations for our experiments, which have been reported to achieve state-of-the-art performances. For our experiments with words, we use \textsc{Muse} representations \cite{Conneau2018} and for our sentence-based experiments, we use \textsc{Laser} representations \cite{Artetxe2018}. Their architectures are described in section \ref{s:relatedwork_models}.

\subsubsection{Word-based Model} The \textsc{Muse} representations are available for 29 languages, aligned in a single vector space. For our experiments, we ignore Vietnamese because spot checks indicated quality issues. For all other languages (see Appendix A for a complete list), we load a vocabulary of 200,000 words. The model encodes every word as a 300-dimensional vector.

\subsubsection{Sentence-based Model} The \textsc{Laser} model generates a sentence representation as a list of tokens. Each token is assigned a vector representation that reflects the contexts in which it occurs. We are using the pre-trained multilingual model which is available for 93 languages. The model encodes every sentence as a 1,024-dimensional vector independent of the length of the sentence which facilitates the comparison across sentences.

 \subsection{Comparing Representational Spaces}
 Comparing the relations in representational spaces is an interdisciplinary problem that is routed in linear algebra and has applications in a large range of research areas. In natural language processing, most work on comparing monolingual representational spaces targets the goal of building better multilingual representations. Canonical correlation analysis \cite{Ammar2016,Faruqui2014}, Kullback-Leibler divergence \cite{Dou2018} and Procrustes alignment \cite{Conneau2018} are only a few methods to maximize the similarity between two representational spaces. Recently, similar methods are being used to compare the hidden representations in different neural models \cite{Raghu2017}. In this article, we apply a method that has been introduced to compare representations obtained from computational models with neuroimaging data of human brain activations \cite{Kriegeskorte2008}. For this method, the representational relations are first evaluated for each modality individually in a representational similarity matrix using common similarity measures such as Euclidean, Cosine or Mahalanobis. In a second step, the two matrices are compared with each other using Spearman correlation to analyze whether the identified relations are similar for the two representational modalities. Representational similarity analysis can also be used to compare different modalities of a computational model \cite{abnar2019blackbox}. In our case, a modality refers to a language. In the following, we formally describe the method and introduce the terminology used for the remainder of the article. For simplicity, we focus on words as the unit of analysis, but the same methodology is used for analyzing sentences. 
 
\paragraph{Similarity vector for a word}
For every English word in our word list of size $N$, we obtain the vector $\mathbf{w_i}$ from our model. We then define the similarity vector $\mathbf{\hat{w}_i}$ for a word vector $\mathbf{w_i}$ such that every element $\hat{w}_{ij}$ of the vector is determined by the cosine similarity between $\mathbf{w_i}$ and the vector $\mathbf{w_j}$ for the j-th word in our word list:
\begin{equation}\label{eq:monolingualSim}
\mathbf{\hat{w}_i} =(\hat{w}_{i1}, \hat{w}_{i2}, \ldots, \hat{w}_{iN}); \quad \hat{w}_{ij} := \cos(\theta_{{w_i},{w_j}})
\end{equation}

For example, if our list consists of the words (\textit{dog, cat, house}), the similarity vector for \textit{cat} would be: 
$\mathbf{\hat{w}_{cat}}$ = $(\cos(\theta_{w_{cat},w_{dog}}),\cos(\theta_{w_{cat},w_{cat}}) ,\cos(\theta_{w_{cat},w_{house}}))$\\

The symmetric matrix consisting of all similarity vectors is commonly referred to as the representational similarity matrix.\footnote{\citet{Kriegeskorte2008} used the term representational dissimilarity matrix (RDM) in the original paper because they measured the distance between representations (the inverse of similarity).} In this example, it would be a matrix with three rows and three columns. 

Note, that the similarity vector is comparable to the similarity vector by \citet{Hamilton2016} which is used to measure semantic drift over time. In our case, the group of "neighbors" to analyze is set to the words in our list to ensure cross-lingual comparability. The underlying concept is also comparable to semantic network analyses \cite{Li2019,Espana2018}.  

\paragraph{Translation of a word}
In order to extract the representational similarity matrix for other languages, we first need to obtain translations for all words $w_i$ in our lists. We do not rely on external translation resources and directly use the information in the multilingual representations instead. We determine the translation $\mathbf{v_i}$ of an English vector $\mathbf{w_i}$ into another language $V$ as its nearest neighbor $\mathbf{v'}$ in the semantic space of the target language:
\begin{equation}\label{eq:trans}
 \mathbf{v_i} := \argmax_{\mathbf{v'} \in V}\,[\cos(\theta_{\mathbf{w_i},\mathbf{v'}})] \quad %\textnormal{for } 1 \le i \le \lvert \mathbb{W} \rvert,  1 \le k \le \lvert V \rvert
\end{equation}

The Spanish translation of $\mathbf{w_{dog}}$ would thus be the vector $\mathbf{v_{perro}}$ assuming that the Spanish word \textit{perro} is the nearest neighbour of $\mathbf{w_{dog}}$ in our model for the Spanish vocabulary. Based on the extracted translations, we can calculate the representational similarity matrices for each language. We then build a second-order matrix to compare the similarity across languages. 

\paragraph{Similarity of two languages}
We can determine the similarity between $\mathbf{w_i}$ and its translation $\mathbf{v_i}$ as the Spearman correlation $\rho$ of their similarity vectors.
\begin{equation}\label{eq:wordSim}
\textnormal{sim}(\mathbf{w_i},\mathbf{v_i}) := \rho(\mathbf{\hat{w}_i}, \mathbf{\hat{v}_i})
\end{equation}
This is comparable to the local neighborhood measure by \citet{Hamilton2016}, but they use cosine distance instead.\footnote{We use Spearman correlation because it is recommended for representational similarity analysis \citet{Kriegeskorte2008} and is also used in \citet{Hamilton2016diachronic}.} This measure can be generalized to express the similarity between the two languages $W$ and $V$ by taking the mean over all $N$ words in our list. 
\begin{equation}\label{eq:crosslingualSim}
\textnormal{sim}(W, V) = \frac{\sum\limits_{i=1}^{N}\rho(\mathbf{\hat{w}_i}, \mathbf{\hat{v}_i)}}{N}
\end{equation}
This definition can easily be extended to any pair of languages. In this case, both similarity vectors are calculated over the corresponding translations in each language. The second-order similarity matrix contains the similarity values for all possible pairs of languages. Our two-step approach has the advantage that it would even work with monolingual representations, if the translations had been obtained from another resource, such as the database \textsc{NorthEuraLex} \cite{dellert2017northeuralex}. Such a resource-driven approach can provide a more accurate source when analyzing finegrained linguistic hypotheses. In this research, we exploit the translation relations inherent in the multilingual model to analyze whether this data-driven approach also captures phenomena of semantic drift. 

 \paragraph{Phylogenetic reconstruction}
Based on the second-order similarity matrix calculated over all languages, we can identify relations between languages. Similarly to \citet{Rabinovich2017}, we perform hierarchical
language clustering using Ward's variance minimization algorithm \cite{Ward1963} as a linkage method to attempt phylogenetic reconstruction. Ward's method iteratively minimizes the total within-cluster variance by, at each step, using this objective as a criterion for selecting new pairs of clusters to merge. To measure the distance between data points, Euclidean distance is used. 
Whereas \citet{Rabinovich2017} use a large set of features as input, we only use the similarity value described above. This value captures to which extent semantic relations between words follow similar patterns in the two languages. 

\paragraph{Tree evaluation} Phylogenetic reconstruction approaches and in particular the evaluation of generated trees are heatedly debated topics and there does not yet exist a standardized procedure \cite{Ringe2002}. Quantitative evaluations thus need to be interpreted very carefully. \citet{Rabinovich2017} propose to evaluate the generated tree with respect to a so-called "gold tree" (see Figure \ref{fig:goldtree}) which was developed by \citet{Serva2008}. \citet{Rabinovich2017} concede that this gold tree has also been questioned and that linguistic researchers have not yet converged on a commonly accepted tree of the Indo-European languages. However, the debatable cases involve more fine-grained distinctions than the ones under analysis here. In case of doubt, we consulted \textsc{Glottolog} as additional reference \cite{Glottolog}. For our quantitative comparison in section \ref{s:quantitative}, we follow the proposed evaluation method and calculate the distance of a generated tree $t$ to the gold tree $g$ by summing over all possible pairs $(W,V)$ of the $M$ leaves (in our case, leaves are languages). For each pair, the difference between the distance $D$ of $W$ and $V$ in the gold tree and the generated tree is squared. $D$ is calculated as the number of edges between $W$ and $V$. 
\begin{equation}\label{eq:distantscore}
Dist(t, g) = \sum_{W,V\in[1,M];W\neq V}(D_t(W,V)-D_{g}(W,V))^2
\end{equation}

As the distance score is dependent on the number of leaves of a tree, we compare the result to reasonable baselines (see Section \ref{s:quantitative}). Our code is available at \url{https://github.com/beinborn/SemanticDrift} to make our modelling decisions more transparent and all experiments reproducible.

%\footnote{\citet{Rabinovich2017} describe that they normalize the distance by the score for a maximally different tree, but they do not provide the value for this maximum. The detected maximum was a distance of 4,520.  Note that our results would improve, if the true maximum is higher than our detected value. Our estimate can thus be considered as a lower bound on the quality. 
  
\paragraph{Multilingual semantic drift}
For detecting diachronic semantic drift, the comparison of similarity vectors of the same word obtained from corpora spanning different decades can easily be interpreted as a time series \cite{Hamilton2016diachronic} or as a function over time \cite{Rosenfeld2018}. For multilingual analyses, an ordering of the languages is not possible because they are all constantly evolving. We thus propose to analyze semantic drift between language families. We postulate that for words which undergo significant semantic drift across languages, the semantic relations are highly correlated within the language family and less correlated outside the family. We assume that the languages are grouped into mutually exclusive sets ${C_j}$ that are chosen based on a research hypothesis. We refer to these sets as clusters $\mathbb{C}$.\footnote{Note, that we leave it unspecified here how the clusters are determined. They can be formed either based on theory-driven knowledge of language families or by empirical observation of language relatedness (for example, according to the results of the phylogenetic reconstruction).}
We iterate through all possible pairs of languages $(W,V); W \in C_j, V \in C_k, W \neq V$ and calculate the Spearman correlation $\rho$ for the respective similarity vectors $\hat{w}_i$ and $\hat{v}_i$.
We define the list of intra-cluster similarities ICS for the i-th word to be the Spearman correlation $\rho$ of the two similarity vectors $\hat{w}_i$ and $\hat{v}_i$ for all pairs that are members of the same cluster ($C_j = C_k$). Accordingly, we determine the cross-cluster similarities CCS for all possible pairs that are in different clusters ($C_j \neq C_k$). 
\begin{equation} \label{eq:ics}
\begin{split} 
\textnormal{ICS}_i &:= \{\rho(\mathbf{\hat{w}_i}, \mathbf{\hat{v}_i}) \mid C_j = C_k\}\\
\textnormal{CCS}_i &:= \{\rho(\mathbf{\hat{w}_i}, \mathbf{\hat{v}_i}) \mid C_j \neq C_k\}\\
\end{split} 
\end{equation}

To calculate the semantic drift for the i-th word over the set of clusters $\mathbb{C}$, we subtract the mean of all cross-cluster similarities from the mean of all intra-cluster similarities. Note, that the value for semantic drift can also be negative if the clusters are not well chosen and the similarity outside clusters is higher than inside clusters.

\begin{equation}\label{eq:drift}
\textnormal{Semantic drift}(i, \mathbb{C}) = 
\frac{\sum \textnormal{ICS}_i}{\lvert \textnormal{ICS}_i \rvert} \; - \frac{\sum \textnormal{CCS}_i}{\lvert \textnormal{CCS}_i \rvert}
\end{equation}

Consider the following simple example with two clusters ($\mathbb{C} = {(es, pt), (de, nl)}$) and the word \textit{dog}.  
The semantic drift is calculated as the mean Spearman correlation of the similarity vectors for the language pairs (es,~pt) and (de,~nl) minus the mean Spearman correlation for all other possible pairs. 

\begin{equation*}
\begin{split}
    \textnormal{drift(dog, } \mathbb{C})=& \textnormal{ mean}(\rho({es}_{\textit{dog}},{pt}_{\textit{dog}}),\rho({de}_{\textit{dog}}, {nl}_{\textit{dog}}))\\
    &- \; \textnormal{mean}(\rho({es}_{\textit{dog}}, {de}_{\textit{dog}}), \rho({es}_{\textit{dog}} ,{nl}_{\textit{dog}}), \rho({pt}_{\textit{dog}}, {de}_{\textit{dog}}), \rho({pt}_{\textit{dog}}, {nl}_{\textit{dog}}))
    \end{split}
    \end{equation*}

We apply our methodology for a series of experiments. We first estimate the quality of the multilingual models for our datasets and then present results for representational similarity analysis and language clustering. 

\begin{table}
\addtolength{\tabcolsep}{-2pt}
\centering
\begin{tabular}{lrrrrrrrrrrrrrrrr} 

\toprule
 & es & it & de & pt & fr & nl & id & ca & pl & no & ro & ru & da & cs & fi \\
\midrule
\rowcolor{black!5}Pereira & 97 & 94 & 93  & 93 &92 & 91 & 88 & 88 & 88 & 87 & 87 & 86 & 85 & 85  &85\\
\rowcolor{black!5}Swadesh & 92 & 92 &88  & 88 &87 &88 & 86 & 81 & 72 & 79 & 75 & 80 & 77 & 76 & 74 \\
\midrule
Europarl & 100 &  99& 100 & 99 &100  & 99  &--  & -- & 99 & -- & 100  & -- & 100 & 100 & -- \\%[1ex]
\\
\hdashline 
\\
& hr & uk & sv & he & hu & bg & tr & el & mk  &sl &et & sk & lt & lv & \textbf{Avg}\\
\midrule
 \rowcolor{black!5}Pereira  &  84 & 82 & 82 & 82 & 82 & 81 & 81 & 80 & 79 & 78 & 77 & 75 & -- &-- & \textbf{85} \\
 \rowcolor{black!5}Swadesh & 80 & 75 & 72 & 71 & 70 & 78 & 69 & 69  & 67 & 73 & 67  & 62 & --& -- &\textbf{77} \\
\midrule
Europarl &--  & -- &99 & --  & -- & 100 & -- & --  & --& 100  &--  &100  &100 & 100 &\textbf{100}\\[1ex]
\bottomrule
\end{tabular}
\caption{Translation quality of the multilingual models (in \%) evaluated by using a dictionary look-up in two resources for the word-based model (rows 1 and 2) and by using similarity search for the sentence-based model (row 3).}
\label{tbl:translationQuality}
\end{table}
\section{Quality of Multilingual Representations}
The quality of monolingual word representations is commonly estimated by evaluating to which extent words that are semantically similar (such as \textit{lake} and \textit{pond}) can be found in close proximity of each other in the semantic space. For multilingual models, the goal is to minimize the representational distance between words which are translations of each other. We check the model quality for a stimulus by determining the nearest neighbor of the stimulus for each target language (see Equation \ref{eq:trans}) and comparing it to a list of known translations. 

The interpretation of the semantic space of sentence vectors is more complex because we can generate infinitely many possible sentences due to the compositionality of language. As a consequence, it is hard to define which sentences should be present in the neighborhood of a sentence even in monolingual space. A sentence with synonyms? The same sentence in another tense? The negated form of the sentence? When we are moving to multilingual space, the monolingual constraint remains fuzzy, but the multilingual training objective is clear: sentences that are aligned as translations in parallel corpora should be close to each other. 

The results for both word-based and sentence-based translation quality assessments are reported in Table \ref{tbl:translationQuality} and discussed in more detail below (see Appendix A and B for a list of languages and their abbreviations). 

\subsection{Translation quality for word-based model}
As the stimuli are presented without context, they might be translated with respect to any of their senses. In order to account for this wide range of translations, we use the multilingual resource \textsc{Babelnet}  \cite{Navigli2010}. For each stimulus word, we collect the translations of all senses of all possible synsets and check whether any of these translations matches the nearest neighbor in the target language retrieved by the \textsc{Muse} model. As the words in the model are not lemmatized, we additionally check for close matches using the \textit{difflib}-module in Python.\footnote{A better approach would be to use language-specific lemmatizers, but they are not available for all languages.} We also count a match if the nearest neighbor of the word is the word itself used as a loanword in the target language. We noticed that the coverage of \textsc{Babelnet} is insufficient for the \textsc{Swadesh} stimuli because it does not contain simple words like \textit{you} or \textit{with}. To account for this, we additionally consult ground-truth dictionaries.\footnote{Available at \url{https://github.com/facebookresearch/MUSE\#ground-truth-bilingual-dictionaries, last accessed: July 1, 2019}}  

\paragraph{Results}
The translation quality for the \textsc{Muse} model is not perfect, but higher than reported in \citet{Artetxe2018} because we use a smaller set of stimuli. The quality is better for the \textsc{Pereira} stimuli (row 1) than for the highly frequent \textsc{Swadesh} stimuli (row 2). As frequency correlates with polysemy, the translation options for these stimuli might be more fuzzy. We had a closer look at the incorrect entries for some languages and noted that the nearest neighbor usually points to semantically highly related words. For example, the nearest German neighbor of the stimulus \textit{silly} is \textit{l\"{a}cherlich} which means ridiculous in English.
We observe that the translation quality tends to decrease for languages that are less similar to English. This indicates that the model provides an interesting testbed for examining semantic drift movements across languages.

\subsection{Translation quality for sentence-based model}
For the sentence-based experiments, we count a perfect match if the nearest neighbor of a sentence in the target space matches the translation of the sentence in the corpus. \citet{Schwenk2017} refer to this method as similarity search. We find that the quality is almost flawless independent of the sentence length. The results indicate that semantic drift phenomena are more likely to occur in the word-based model because the sentence-based model exhibits less variation across languages. It is optimized with respect to multilingual translation, whereas the word-based model balances monolingual semantic similarity and cross-lingual translation constraints and optimizes them jointly. 

% with 198-200 perfect matches (out of 200 sentences) for all languages. Only the quality for Swedish is a little lower for the short sentences (196 perfect matches).

\section{Representational Similarity}
In this section, we illustrate the steps of our analysis method. We first look at intralingual semantic relations and then perform a cross-lingual representational similarity analysis. We focus on the word-based model because example words can be visualized more intuitively than full sentences.

\begin{figure}
\centering
  \includegraphics[width=\textwidth]{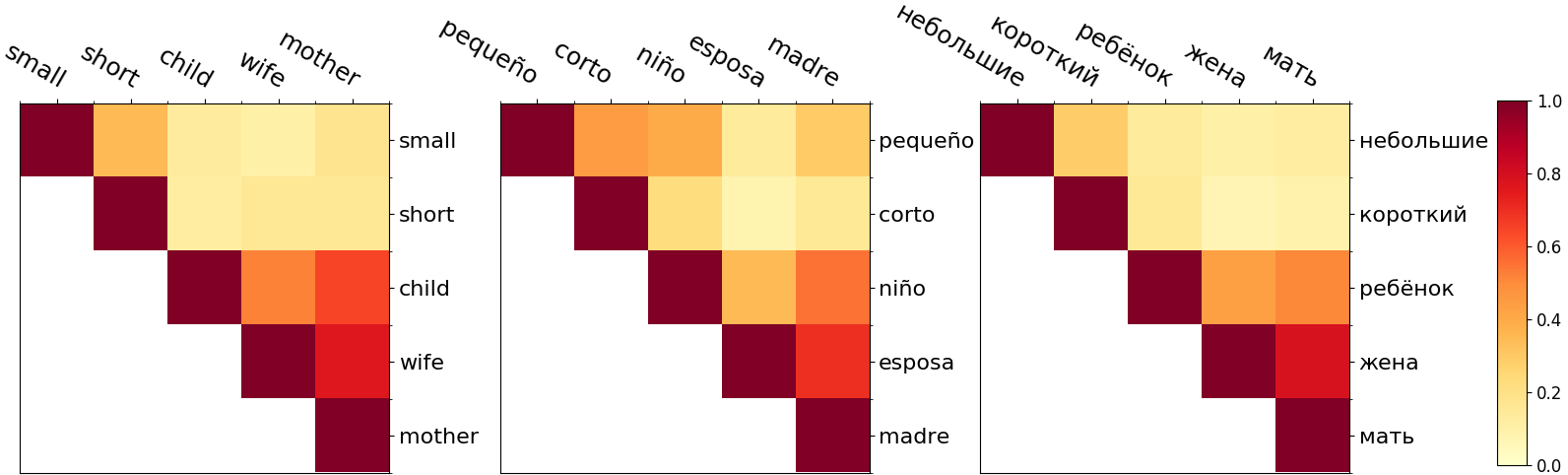}
\caption{Cosine similarity between vector pairs for the English words \textit{small}, \textit{short}, \textit{child}, \textit{wife}, \textit{mother} and for the nearest neighbors of the English words in the
Spanish (middle) and Russian (right) representations. }
\label{fig:RDM}
\end{figure}
\subsection{Intralingual Semantic Relations}
We first extract the English vectors for all words in our list. We analyze the cosine similarity between the vectors and construct a representational similarity matrix as described in Equation \ref{eq:monolingualSim}. We then extract the translations for each word in our lists as described in Equation \ref{eq:trans} to construct representational similarity matrices for all languages. 

Figure \ref{fig:RDM} illustrates example matrices for a subset of the five words \textit{small}, \textit{short}, \textit{child}, \textit{wife}, \textit{mother} for English, Spanish, and Russian. It can be seen that the similarity patterns are comparable, but we also note some differences. For example, the extracted Spanish words \textit{ni\~no} and \textit{peque\~no} are more similar to each other than their translations \textit{child} and \textit{small}. We assume that this is due to the fact that both \textit{small} and \textit{little} are translated as \textit{peque\~no} in Spanish. This illustration indicates that semantic relations vary slightly across languages. Note that the nearest Russian neighbor \textit{\foreignlanguage{russian}{небольшие }}is not the most intuitive translation for \textit{small} because it is a plural form (cosine similarity: 0.67). This effect occurs because the vocabulary in the \textsc{Muse} model is not lemmatized. We observe that the cosine similarity (0.67) is quite similar for alternatives like \textit{\foreignlanguage{russian}{небольшой} }(0.65) and 
\textit{\foreignlanguage{russian}{маленькие }}(0.61). This indicates that it might be reasonable to analyze the top $n$ nearest neighbors and/or only work with lemmas to obtain purer results from a linguistic perspective.

\subsection{Cross-lingual Representational Similarity Analysis}
\begin{figure}
\centering
  \includegraphics[scale=0.13]{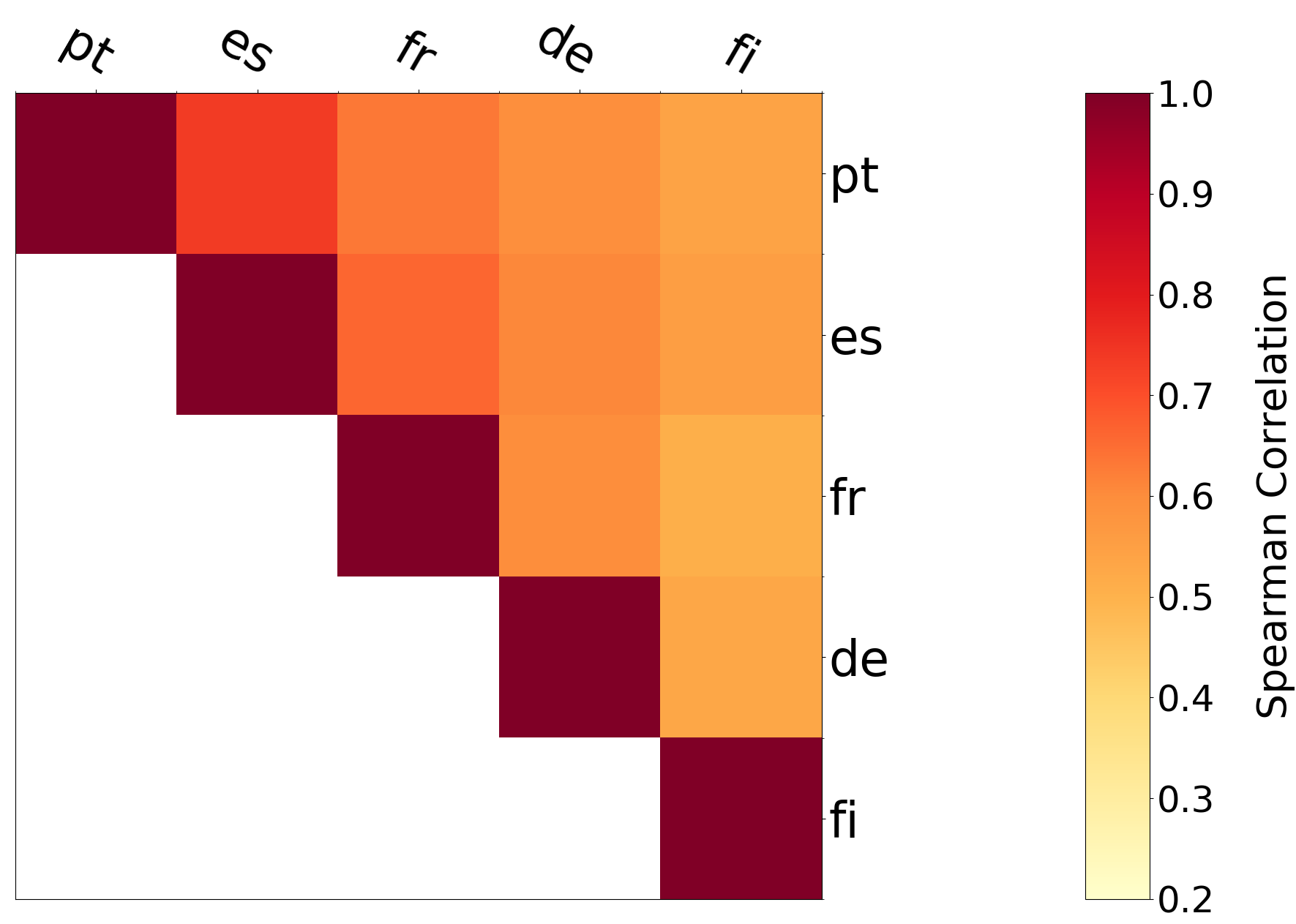}
  \caption{Representational similarity analysis for five selected languages (Portuguese, Spanish, French, German, Finnish). }
  \label{fig:RSA}
\end{figure}

 The intralingual similarity matrices described above serve as the basis for the cross-lingual representational similarity analysis. We measure the correlation of the semantic similarity for each pair of languages as described in Equation \ref{eq:crosslingualSim}. The resulting matrix is illustrated in Figure \ref{fig:RSA} for five selected languages. It can be seen that languages like Spanish (es), Portuguese (pt) and French (fr) have highly similar semantic patterns. For German (de), the similarity is slightly lower and Finnish (fi) stands out as not being very similar to the other languages. 
 
\begin{figure}
\centering
  \includegraphics[scale=0.45]{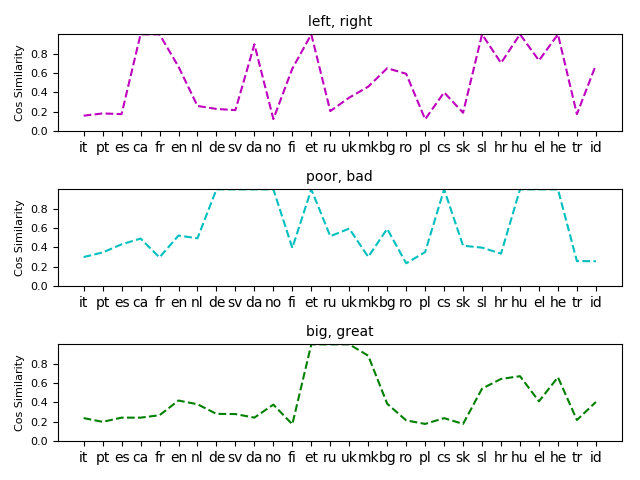}
   \caption{The cosine similarity between two word vectors varies for each language. The plot shows the similarity values for the word pairs with the highest variance across languages. Interestingly, all six words are adjectives.}

  \label{fig:wordpairs}
\end{figure}

The second-order similarity matrix only indicates the average correlation between the similarity vectors of the words. For linguistic analyses, it is more interesting to look at the behavior of individual words and word pairs. In Figure \ref{fig:wordpairs}, we plot the word pairs with the highest variance in similarity across languages. It is interesting to see that all six words are adjectives. Word representations are mostly analyzed on nouns \cite{Finkelstein2002} and sometimes on verbs \cite{Gerz2016}. \citet{Faruqui2014} discuss that separating synonyms and antonyms in word representations can be tricky because they tend to occur in very similar contexts. We find that the nearest French neighbour of both \textit{left} and \textit{right} is \textit{gauche} meaning left. The same phenomenon occurs for Catalan. For Slovenian, both words are translated to \textit{desno} meaning right. For the pairs \textit{big-great} and \textit{poor-bad}, we observe that they are translated to the same word in some languages which is not surprising as they are likely to occur in similar contexts. However, the nearest neighbor of \textit{big} is \textit{big} in many languages because it is often used as a loanword. Unfortunately, the loan word exhibits different semantic properties because it is only used in specific contexts (e.g., products or other named entities), whereas a translation would be used in more common contexts. This explains the low similarity between \textit{big} and \textit{great} for many languages. Our findings indicate that the methodology we introduce for analyzing cross-lingual relations can also be used to identify flaws of the computational model. 

In these three examples, it seems as if the cosine similarities are generally higher for Estonian, Hungarian, Greek, and Hebrew, whereas they are constantly low for Italian, Portuguese, and Spanish. In order to verify whether this observation points to a systematic pattern, we checked the mean and variance scores for the pairwise similarities, but the scores were comparable for all seven languages. 

\section{Language Clustering}
\begin{figure}
\begin{subfigure}{.99\textwidth}
  \includegraphics[scale=0.27]{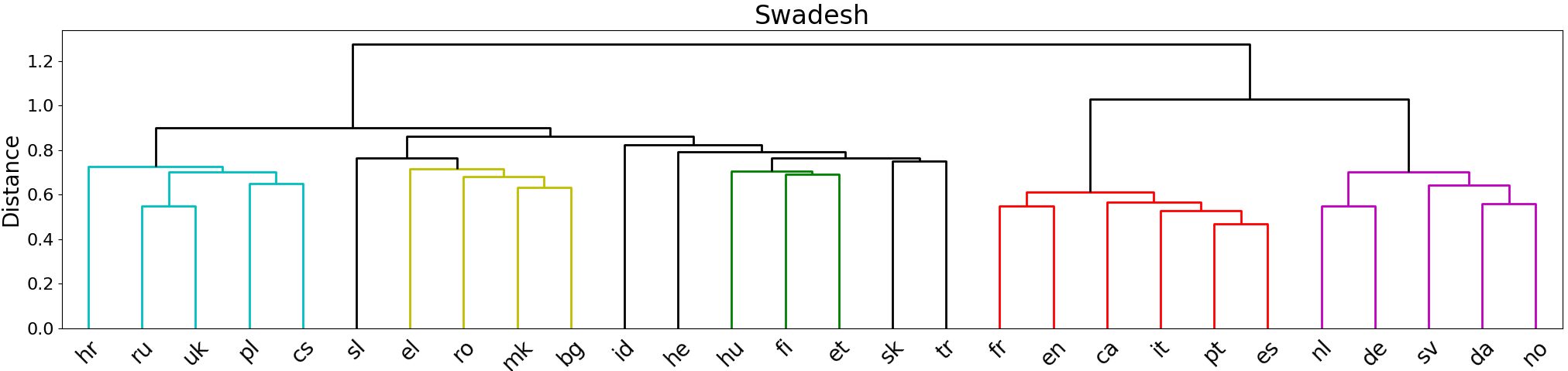}
  \end{subfigure}
\begin{subfigure}{.99\textwidth}
  \includegraphics[scale=0.27]{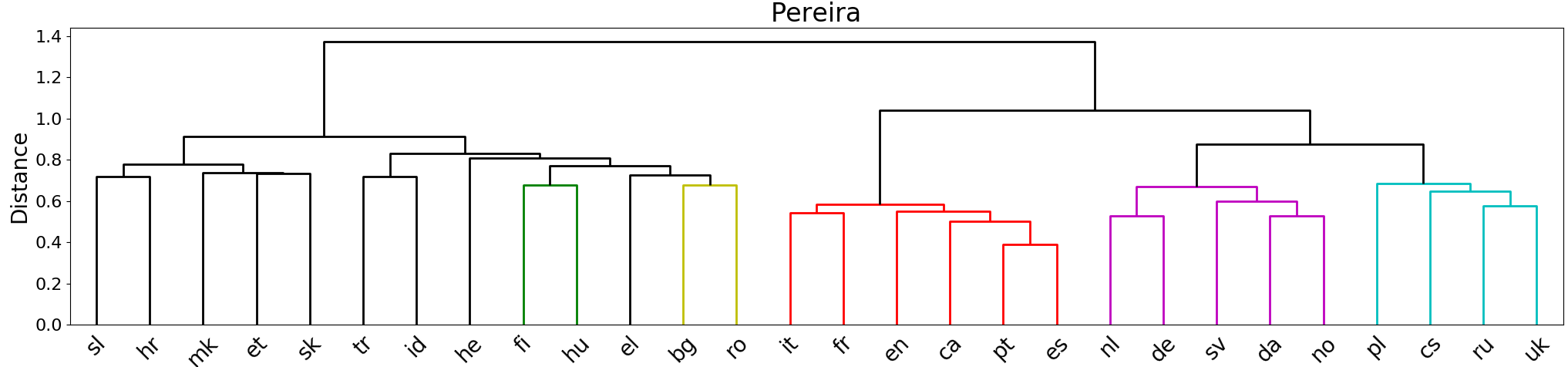}
\end{subfigure}
 \caption{Hierarchical clustering of languages based on the results of the cross-lingual representational similarity analysis of the \textsc{Swadesh} and the \textsc{Pereira} words. }
  \label{fig:Clusterwords}
\end{figure}
We use the result of the representational analysis to run a hierarchical clustering algorithm over the languages. The clustering is only based on the semantic similarity scores for pairs of languages (see section \ref{s:methodology} for details). We first discuss the word-based and the sentence-based results separately and then perform a quantitative evaluation. 

\subsection{Clustering by word-based representational similarity} Surprisingly, our computationally generated trees in Figure \ref{fig:Clusterwords} resemble the trees that are commonly accepted by linguistic experts quite closely. We cross-check our observations against the renowned linguistic resource \textsc{Glottolog} \cite{Glottolog} and observe a clear distinction between Western and Eastern European languages in the generated tree. It is even possible to identify a distinction between Germanic and Latin languages (with the exception of English). Obviously, the extracted cluster tree is not perfect, though. For example, Indonesian (id), Hebrew (he), and Turkish (tr) do not fit well with the rest and Romanian (ro) would be expected a little closer to Italian. 

The subtree containing the languages Russian (ru), Ukranian (uk), Czech (cs), and Polish (pl) is grouped with other Slavic languages for the \textsc{Swadesh} tree and with the Germanic languages in the \textsc{Pereira} tree. This might be an artifact of our quite diverse set of languages spanning many language families including non-Indo-European ones like Finnish (fi), Hebrew (el) and Indonesian (id). Czech and German, for example, are quite related and share many cognates due to historical reasons, so that their closeness in the tree is explainable. The tree using the combined stimuli is more similar to the \textsc{Swadesh} version (see Appendix E, Figure \ref{fig:dendrogram_combined}).  

Furthermore, it is interesting to note that similarly to the tree by \citet{Rabinovich2017}, Romanian and Bulgarian (bg) are clustered together in our trees although they represent different language families (Romance and Slavic languages). Our observations indicate that language contact might be more relevant for semantic drift phenomena than a common ancestor language. The same argument could explain the vicinity of English (en) and French (fr). Our findings support the results by \citet{eger2016language} and \citet{Thompson2018quantifying} which showed that semantic similarity between languages correlates with their cultural and geographic proximity. 

\begin{figure}
\begin{subfigure}{.99\textwidth}
\centering
  \includegraphics[width=13cm]{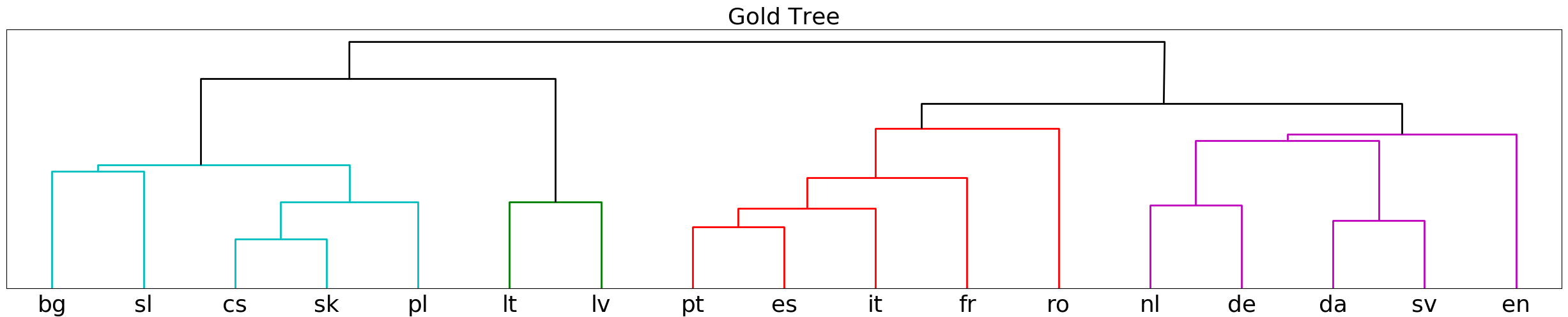}
  \caption{The "gold tree" of the 17 Indo-European languages  used in the sentence-based experiment. It is a pruned version of the tree in \citet{Serva2008}. }
  \label{fig:goldtree}
\end{subfigure}
\begin{subfigure}{.99\textwidth}
\centering
 \includegraphics[width=13.4cm, height=2.9cm]{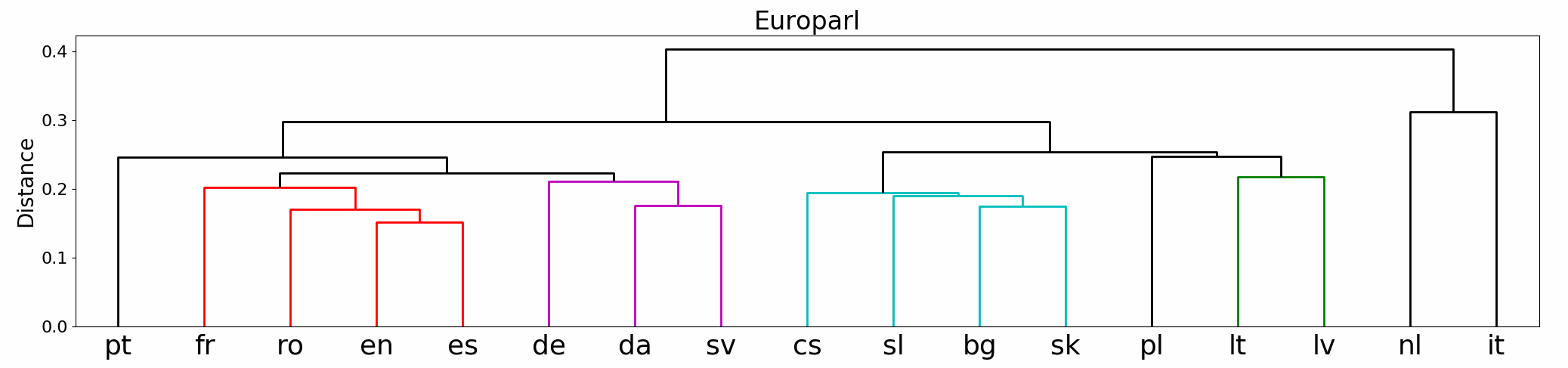}
 \caption{Hierarchical clustering of the 17 languages based on the results of the cross-lingual representational similarity analysis of the \textit{mid} Europarl sentences. }
 \label{fig:ClusterEuroparl}
    \end{subfigure}
\label{fig:sentencebased_clustering}
\caption{The results of the sentence-based language clustering (b) compared to the gold tree (a).}
\end{figure}

\subsection{Clustering by sentence-based representational similarity}
Figure \ref{fig:ClusterEuroparl} shows the results of the clustering using the sentence-based model. We see that the separation of Eastern and Western European languages works quite well, but the more finegrained relations between languages are less accurately clustered than for the word-based experiments. In particular, Dutch (nl) and Italian (it) should be noted as outliers. From a quantitative perspective, we find that the distances between languages (visualized on the y-axis) are much lower than for the word-based experiments.\footnote{See also the correlation values in the representational similarity matrix in Appendix E, Figure \ref{fig:RSAEuroparl}.} Recall that the \textsc{Laser} architecture is optimized for translating between multiple languages. Based on this training objective, it is plausible that subtle differences between languages tend to be smoothed out. In contrast, the word-based \textsc{Muse} model has explicitly been optimize to fulfill both the monolingual objective (preserve intralingual semantic relations) and the cross-lingual objective (representational similarity of translations). 

\subsection{Quantitative evaluation}\label{s:quantitative}
\begin{figure}
\begin{subfigure}{.99\textwidth}
\centering
  \includegraphics[width=13cm,height=2.8cm]{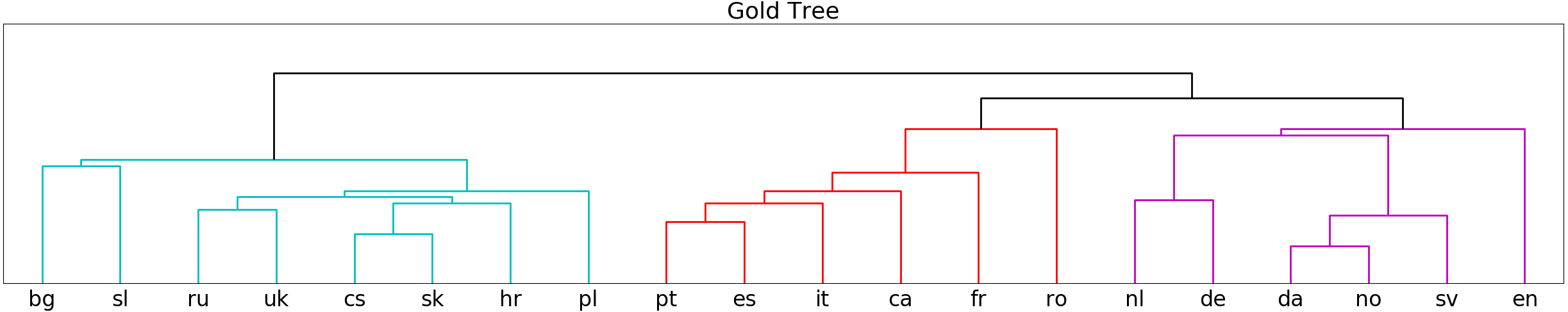}
  \caption{The "gold tree" for the 20 Indo-European languages used in the word-based experiments. It is a pruned version of the tree in \citet{Serva2008}. }
  \label{fig:goldtree_words}
\end{subfigure}
\begin{subfigure}{.99\textwidth}
\centering
 \includegraphics[width=13.6cm,height=3.1cm]{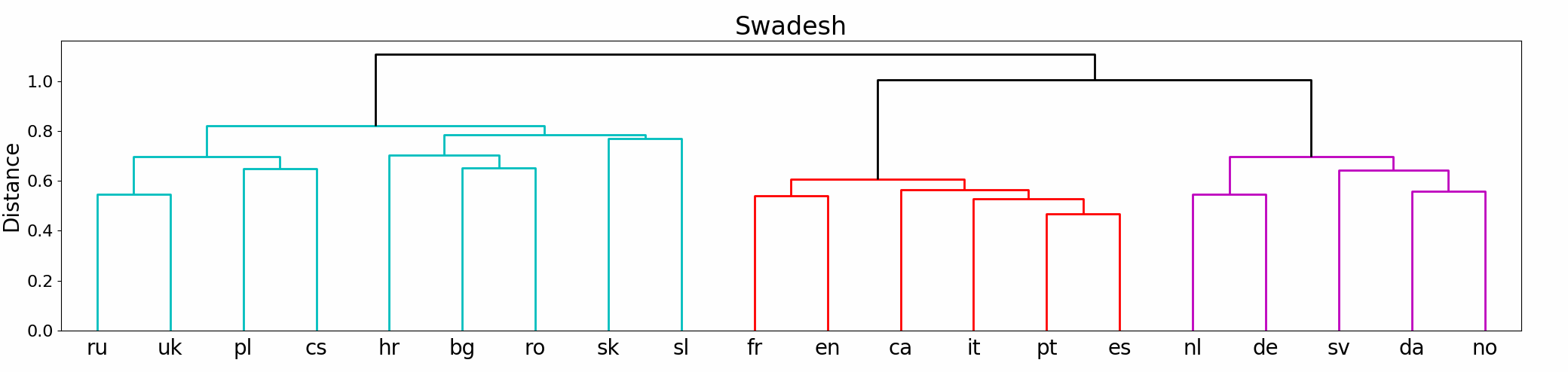}
 \caption{Hierarchical clustering of the 20 languages based on the results of the cross-lingual representational similarity analysis of the \textsc{Swadesh} words.}
 \label{fig:ClusterSwadesh}
    \end{subfigure}
\label{fig:wordbased_clustering}
\caption{The results of the word-based language clustering for a subset of 20 Indo-European languages (b) compared to the gold tree (a).}
\end{figure}

In order to better judge the quality of the language tree, we additionally perform quantitative experiments and compare the distance to a gold tree. For the 17 languages in the sentence-based model, we use the same tree as \citet{Rabinovich2017} which was developed by \citet{Serva2008} (see Figure \ref{fig:goldtree}). For the word-based model, we reduce the set of languages to 20 Indo-European ones and adjust the gold tree accordingly (see Figure \ref{fig:goldtree_words}). We calculate the distance between our trees based on the representational similarity analysis (see Figures \ref{fig:ClusterEuroparl} and \ref{fig:ClusterSwadesh}) to the gold trees as described in Equation \ref{eq:distantscore}. 

As the distance score depends on the number of leaves, the results are not directly comparable for word-based and sentence-based experiments. We thus calculate the average distance score for a randomly generated tree over 50,000 iterations and report the change in quality of our model with respect to that baseline in Table \ref{tbl:treeQuality}. For this random baseline, instead of calculating the similarity matrix of languages as described in section 3.3, we generate it randomly from a uniform distribution between 0.3 and 0.8 (we omit the extreme values because they are unlikely to be observed experimentally).\footnote{We found experimentally that using the full range from 0 to 1 does not make a difference when the number of iterations is high enough. Restricting the range leads to a more rigorous and more plausible baseline.}  This might not be the most plausible distribution, so we calculated an additional permutation baseline. We randomly permute the values of the similarity matrix by our model and average the results over 50,000 iterations.\footnote{We make sure that the scrambled matrix remains symmetric, see code for details.} 

\begin{table}
\caption{Quality changes (in \%) of reconstructed language trees compared to the random baseline. Quality is calculated as distance to the gold tree. As a control condition, we randomly permute the values of our drift model and average the results over 50,000 iterations.}
\label{tbl:treeQuality}
\centering
\begin{tabular}{llrc} 
\toprule
Experiment & Category &  Permutation & Drift Model  \\
\midrule
Words       &   Pereira     &-38.9      & +37.7 \\
            &   Swadesh     &-38.6      & \textbf{+53.4} \\
            &   Combined    &-38.1      &+51.8\\
\midrule
Sentences   &   Short       &\phantom{3}+4.0    &+40.1 \\
            &   Mid         &\phantom{3}+3.4    &\textbf{+44.5}\\
            &   Long        &\phantom{3}+4.0    &+34.0 \\
            \midrule
            &   Rabinovich et al. (2017) &--\phantom{.-}      &+55.5\\
            
\bottomrule
\end{tabular}
\end{table}
\paragraph{Results} We see that the quality of our generated trees is considerably better than chance. Particularly the results for the \textsc{Swadesh} stimuli and the \textsc{mid} sentences stand out as strong improvements over a random baseline. The clustering results seem to negatively correlate with the translation quality of the model (see Table \ref{tbl:translationQuality}): clustering works better for the word-based model than for the sentence-based model and better for \textsc{Swadesh} than for \textsc{Pereira} stimuli. 
We speculate that lower translation quality is a data-driven indicator of semantic distance. As a consequence, the differences between languages become more pronounced in our analysis which leads to better clustering. These findings support our assumption stated in the previous section that models which are optimized for learning universal sentence representations smooth out the differences between language families. From an engineering perspective, this is a reasonable goal, but it might come at a cost of expressivity when it comes to more finegrained linguistic and cultural subtleties.
\footnote{ We noted that if we degrade the homogeneity of the sentence-based model by not applying byte-pair encoding (BPE) on the input the clustering quality improves drastically (on average, 67\% improvement over the random baseline). BPE is used to limit the shared vocabulary of the languages by splitting rare and unknown words into known subword units and it has been shown to improve the results of neural machine translation \cite{Sennrich2016}. We assume that not applying this normalization affects morphologically richer languages more than others and as a consequence increases the variance in the similarity matrix.  }
% A multilingual model like \textsc{Bert} which additionally optimizes the sentence representations to capture whether two sentences are likely to occur as consecutive sentences in a text might be able to better reflect these differences. 

It is interesting to see that the results for the permutation baseline are even worse than for the random baseline in the word-based setting. This shows that our methodology does pick up on the differences between languages. If this inductive bias is scrambled, the clustering results get worse than when treating all languages uniformly. For the sentence-based experiments, we do not observe this effect because the similarity scores are more homogeneous across languages.\footnote{See also Appendix E, Figure \ref{fig:RSAEuroparl}.} 

The results by \citet{Rabinovich2017} are slightly better than our word-based ones because they used hundreds of structural features whereas we only use a single semantic measure. In addition, they tackled a slightly different task as they only worked with English sentences which are translations. It should be noted, that improvements for the sentence-based experiments are easier to obtain because the lower number of languages (17 vs 20) leads to a lower number of possible combinations in the tree. 

\section{Semantic Drift}
The clustering results indicate that the distances between the words vary in the semantic space of different language families. For a qualitative exploration, we have a closer look at semantic drift for three language pairs that are clustered closely together in our trees: 

Spanish and Portuguese (es,~pt) for the Romance languages, German and Dutch (de,~nl) for the Germanic languages, and Russian and Ukranian (ru,~uk) for the Slavic languages. We include this analysis as an illustration of our methodology. The choice of the clusters and their size could be conveniently adjusted for any linguistic hypothesis. 

\begin{figure}
\begin{subfigure}{.52\textwidth}
  \includegraphics[width=\textwidth]{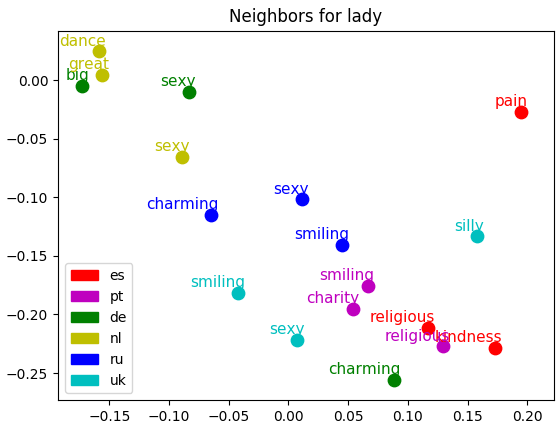}
\end{subfigure}
\begin{subfigure}{.52\textwidth}
  \includegraphics[width=\textwidth]{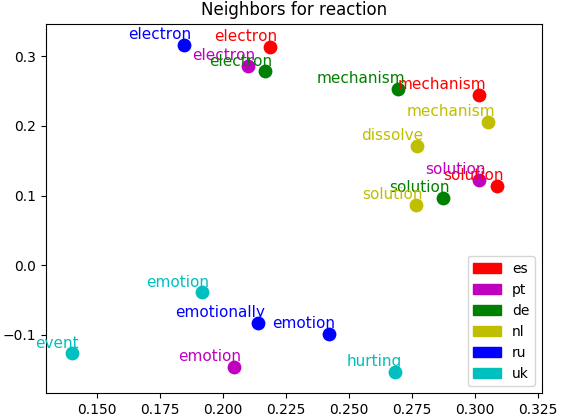}
\end{subfigure}%
  \caption{Examples of words with high Spearman correlation within a cluster and low Spearman correlation outside clusters for three selected clusters (es,~pt), (de,~nl), (ru,~uk). For readability, we always use the English word, but, e.g., \textit{pain} colored in red stands for the nearest Spanish neighbor of \textit{pain} which is \textit{dolor}.}
  \label{fig:semanticDrift}
\end{figure}

For each word in the \textsc{Pereira} list, we calculate the semantic drift as described in Equation \ref{eq:drift}. In Figure \ref{fig:semanticDrift}, we visualize two examples with a high drift score for these clusters. The word representations have been reduced to two dimensions by applying principal components analysis on the joint representational space of all six languages \cite{PCA}. For each language, we plot the \textsc{Pereira} words with the highest cosine similarity to \textit{lady} and \textit{reaction}. For readability, we always use the English word, but, e.g., \textit{pain} colored in red stands for the nearest Spanish neighbor of \textit{pain} which is \textit{dolor}. 

It can be seen that \textit{lady} is close to \textit{religious} for Portuguese and Spanish, but not for the other languages. We note, that the nearest neighbor for \textit{lady} is not its translation, but the loanword itself (or its transliteration) for Dutch, German, and Russian. This explains the similarity to \textit{sexy} and \textit{big} which are also used as loanwords in Dutch and German.
The word \textit{reaction} is a cognate originating from Latin in all six languages. The plot indicates that it is more closely associated with technical terms in the clusters (es,~pt) and (de,~nl), and with emotional terms in the cluster (ru,~uk).

It should be noted, that these examples only serve as anecdotal evidence and that the differences between the languages cannot always be observed when looking at only two dimensions. However, our methodology makes it possible to quantify semantic differences between words across languages. This can be used to better understand flaws of the computational representations (e.g., the observation that words tend to be represented by loanwords even when a more accurate translation exists), and the methodology can also generate hypotheses for historical linguistics when applied on a larger vocabulary. From an application perspective, analyses of semantic drift are particularly interesting for the field of foreign language learning. When understanding a foreign text, learners rely on background knowledge from languages they already know \cite{beinborn2016diss}. Phenomena of semantic drift can thus lead to severe misunderstandings and should receive increased attention in education.  

\section{Discussion}
We have introduced a methodology to analyze semantic drift in multilingual distributional representations. 
Our analyses show that by comparing the representational distances for a test set of about 200 words, we can reconstruct a phylogenetic tree that closely resembles those assumed by linguistic experts. These results indicate that multilingual distributional representations which are only trained on monolingual text and bilingual dictionaries preserve relations between languages without the need for any etymological information. Methods in lexicostatistics have previously been criticized for relying on  subjective cognate judgments \cite{Geisler2010}. A certain level of subjectivity might also be present in the "ground-truth" bilingual dictionaries used for the computational multilingual representations that were analyzed in this article. However, the large vocabulary should help to balance out potential biases. 

So far, multilingual representations have mostly been evaluated based on their performance on specific tasks. In this article, we look beyond engineering goals and analyze the semantic relations between languages in computational representations. We find that the word-based model captures differences in the semantic structure that correspond to linguistic expectations. The sentence-based model, on the other hand, seems to be optimized to balance out subtle differences between language families. This might be a suitable scenario for obtaining better machine translation results, but for linguistic analyses the training objective would have to be adjusted towards maintaining some language diversity. Another important aspect is the training data of the computational models. The corpora used for the word-based model might be less balanced across languages and as a consequence, differences between languages are reinforced. 

For future work, analyses of semantic drift in multilingual representations can serve two main purposes: 
From a technical perspective, they can indicate unwanted characteristics in the multilingual representations and steer processes for technical improvement of multilingual models. In our analyses, we have seen that words tend to be close to identical loanwords in the target space even if a more accurate translation is available. Loanwords often find their way into a language to add nuances to the semantic inventory. As a consequence, they tend to occur only in specific contexts that call for these nuances. The semantic relations to other words can thus be biased due to the introduction of the loanword. In addition, we find that adjectives are not well separated from their antonyms in the semantic space. This indicates that relying on co-occurrence patterns might not be sufficient for capturing semantic relations in word classes other than nouns. 

From a linguistic perspective, our methods provide a quantitative means to study linguistic phenomena across languages. The development of multilingual computational models opens up new possibilities for comparative linguistics. In this paper, we have laid out a methodology to query these models for semantic drift. The results of these queries can be used to generate hypotheses for historical linguistics and social linguistics because they indicate similarities in the organization of semantic concepts.

Our word-based experiments used English as anchor language for obtaining translations. This is not an unreasonable choice as most multilingual computational models have been developed from an English perspective. However, it poses limitations on the interpretation of the linguistic results. For future work, we recommend to take a more multilingual perspective. It should also be noted that our methods cannot capture phonetic or phonological changes such as vowel shift. We understand our proposed methodology as an addition to the inventory of linguistic analysis not as a replacement. 

\section{Conclusion}
We introduced a methodology to analyze the semantic structure of multilingual distributional representations. Our method is inspired by research in neuroscience on comparing computational representations to human brain data. We adapted the analysis to compare representations across language families. We show that our method can be used for phylogenetic reconstruction and that it captures subtle semantic differences of words between language families. In addition, we proposed a new measure for identifying phenomena of semantic drift. Our qualitative examples indicate that this measure can generate new hypotheses for comparative linguistics. 

The computational models for sentences are available for a huge range of languages. In this article, we restricted the languages to those used in previous work for a reasonable comparison. We now plan to corroborate our findings on the whole spectrum and to further extend the word-based analyses of semantic drift.

\begin{acknowledgments}
The work presented here was funded by the Netherlands Organisation for Scientific Research (NWO),
through a Gravitation Grant 024.001.006 to the Language in Interaction Consortium.
We gratefully acknowledge Bas Cornelissen and Tom Lentz for valuable discussions of earlier versions of the article. 
We would like to thank the anonymous reviewers for their very constructive and helpful feedback and their attention to detail. 
 \end{acknowledgments}

\starttwocolumn
\bibliography{compling_style}

\newpage
\onecolumn
\appendix 
% Overlapping words: big, bird, blood, burn, dig, dog, fight, fish, flow, hair, laugh, left, mountain, play, road, sew, skin, star, tree, wash
\appendixsection{Languages for Word-based Experiments}
Bulgarian (bg), Catalan (ca), Croatian (hr), Czech (cs), Danish (da), Dutch (nl), English (en), Estonian (et), Finnish (fi), French (fr), German (de), Greek (el), Hebrew (he), Hungarian (hu), Indonesian (id), Italian (it), Macedonian (mk), Norwegian (no), Polish (pl), Portuguese (pt), Romanian (ro), Russian (ru), Slovakian (sk), Slovenian (sl), Spanish (es), Swedish (sv), Turkish (tr), Ukrainian (uk)
\appendixsection{Languages for Sentence-based experiments}\label{appendix:sentlang}
Bulgarian (bg), Czech (cs), Danish (da), Dutch (nl), English (en), French (fr), German (de), Italian (it), Latvian (lv), Lithuanian (lt), Polish (pl), Portuguese (pt), Romanian (ro), Slovakian (sk), Slovenian (sl), Spanish (es), Swedish (sv)
\appendixsection{Swadesh words}\label{app:swadeshwords}
all, and, animal, ashes, at, back, bad, bark, because, belly, big, bird, bite, black, blood, blow, bone, breast, breathe, burn, child, cloud, cold, come, correct, count, cut, day, die, dig, dirty, dog, drink, dry, dull, dust, ear, earth, eat, egg, eye, fall, far, fat, father, fear, feather, few, fight, fingernail, fire, fish, five, float, flow, flower, fly, fog, foot, forest, four, freeze, fruit, full, give, good, grass, green, guts, hair, hand, he, head, hear, heart, heavy, here, hit, hold, horn, how, hunt, husband, i, ice, if, in, kill, knee, know, lake, laugh, leaf, left, leg, lie, live, liver, long, louse, man, many, meat, moon, mother, mountain, mouth, name, narrow, near, neck, new, night, nose, not, old, one, other, play, pull, push, rain, red, right, river, road, root, rope, rotten, round, rub, salt, sand, say, scratch, sea, see, seed, sew, sharp, short, sing, sit, skin, sky, sleep, small, smell, smoke, smooth, snake, snow, some, spit, split, squeeze, stab, stand, star, stick, stone, straight, suck, sun, swell, swim, tail, that, there, they, thick, thin, think, this, three, throw, tie, tongue, tooth, tree, turn, two, vomit, walk, warm, wash, water, we, wet, what, when, where, white, who, wide, wife, wind, wing, wipe, with, woman, worm, year, yellow, you
\appendixsection{Pereira words} \label{app:pereirawords}
ability, accomplished, angry, apartment, applause, argument, argumentatively, art, attitude, bag, ball, bar, bear, beat, bed, beer, big, bird, blood, body, brain, broken, building, burn, business, camera, carefully, challenge, charity, charming, clothes, cockroach, code, collection, computer, construction, cook, counting, crazy, damage, dance, dangerous, deceive, dedication, deliberately, delivery, dessert, device, dig, dinner, disease, dissolve, disturb, do, doctor, dog, dressing, driver, economy, election, electron, elegance, emotion, emotionally, engine, event, experiment, extremely, feeling, fight, fish, flow, food, garbage, gold, great, gun, hair, help, hurting, ignorance, illness, impress, invention, investigation, invisible, job, jungle, kindness, king, lady, land, laugh, law, left, level, liar, light, magic, marriage, material, mathematical, mechanism, medication, money, mountain, movement, movie, music, nation, news, noise, obligation, pain, personality, philosophy, picture, pig, plan, plant, play, pleasure, poor, prison, professional, protection, quality, reaction, read, relationship, religious, residence, road, sad, science, seafood, sell, sew, sexy, shape, ship, show, sign, silly, sin, skin, smart, smiling, solution, soul, sound, spoke, star, student, stupid, successful, sugar, suspect, table, taste, team, texture, time, tool, toy, tree, trial, tried, typical, unaware, usable, useless, vacation, war, wash, weak, wear, weather, willingly, word

\appendixsection{Additional Figures} \label{app:figures}
\setcounter{figure}{7}
\begin{figure}[h]
\centering
  \includegraphics[width=\textwidth]{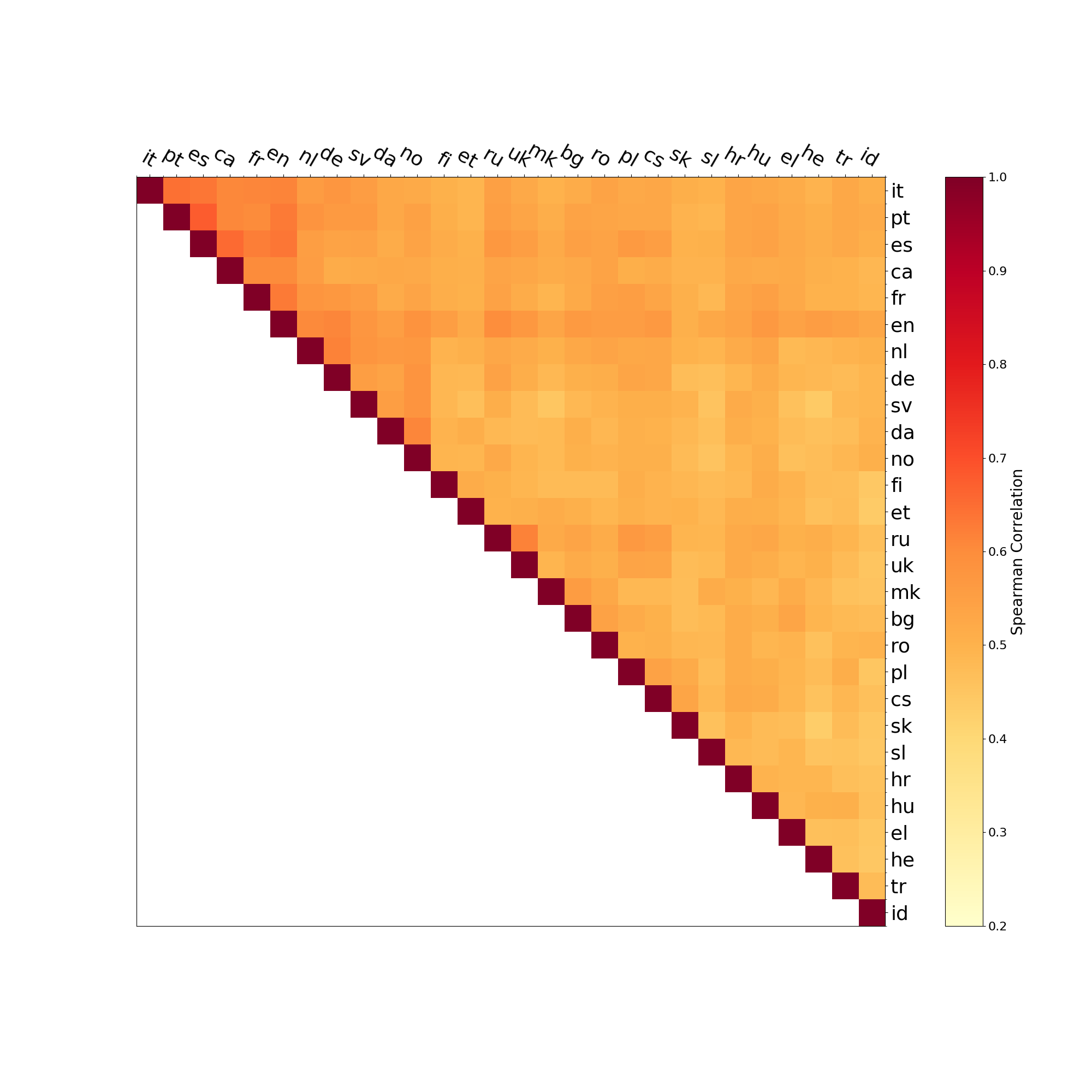}
\caption{Full representational similarity analysis for the \textsc{Swadesh} words.}
\label{fig:RSASwadesh}
\end{figure}
\begin{figure}
\centering
  \includegraphics[width=\textwidth]{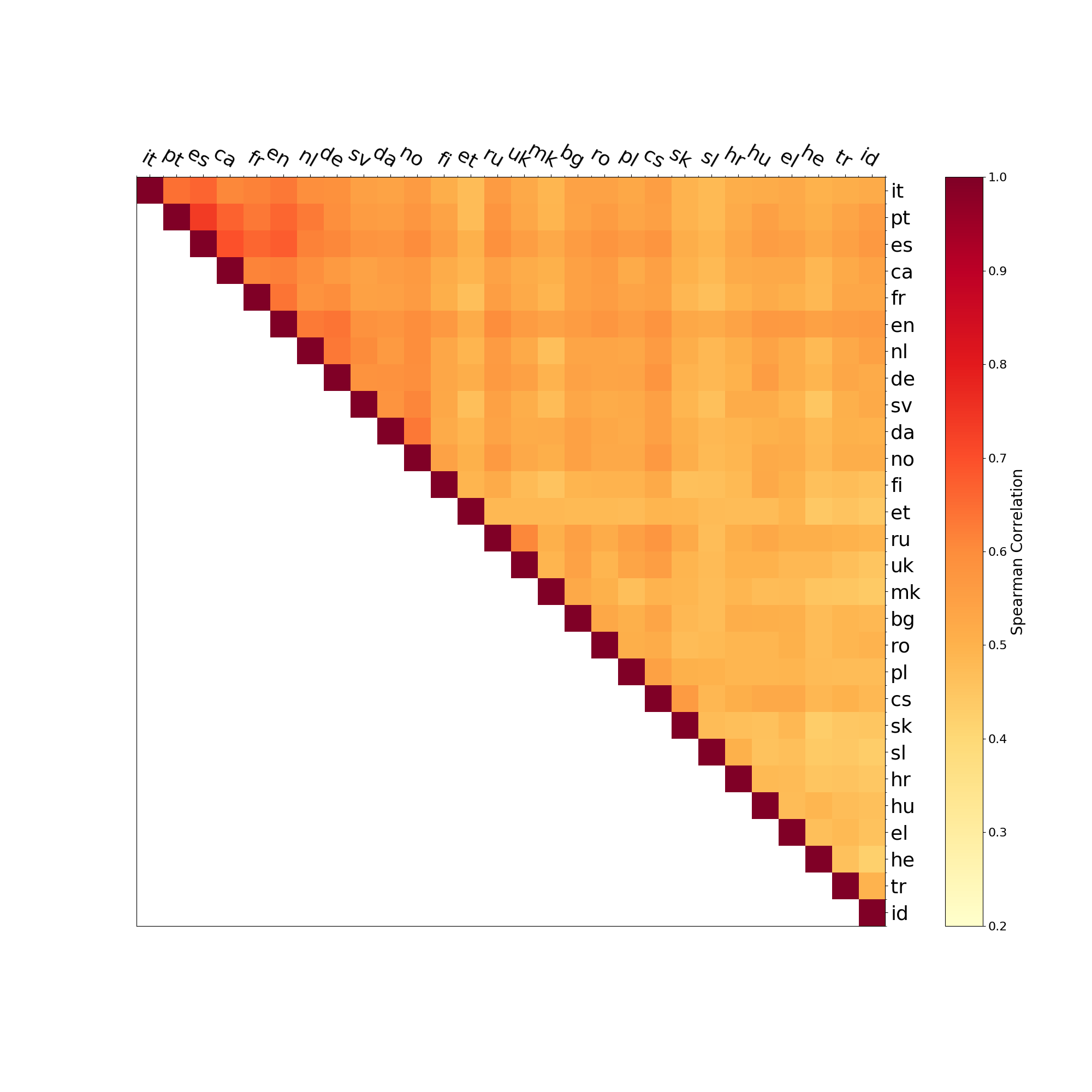}
\caption{Full representational similarity analysis for the \textsc{Pereira} words.}
\label{fig:RSAPereira}
\end{figure}
\begin{figure}
\centering
  \includegraphics[width=\textwidth]{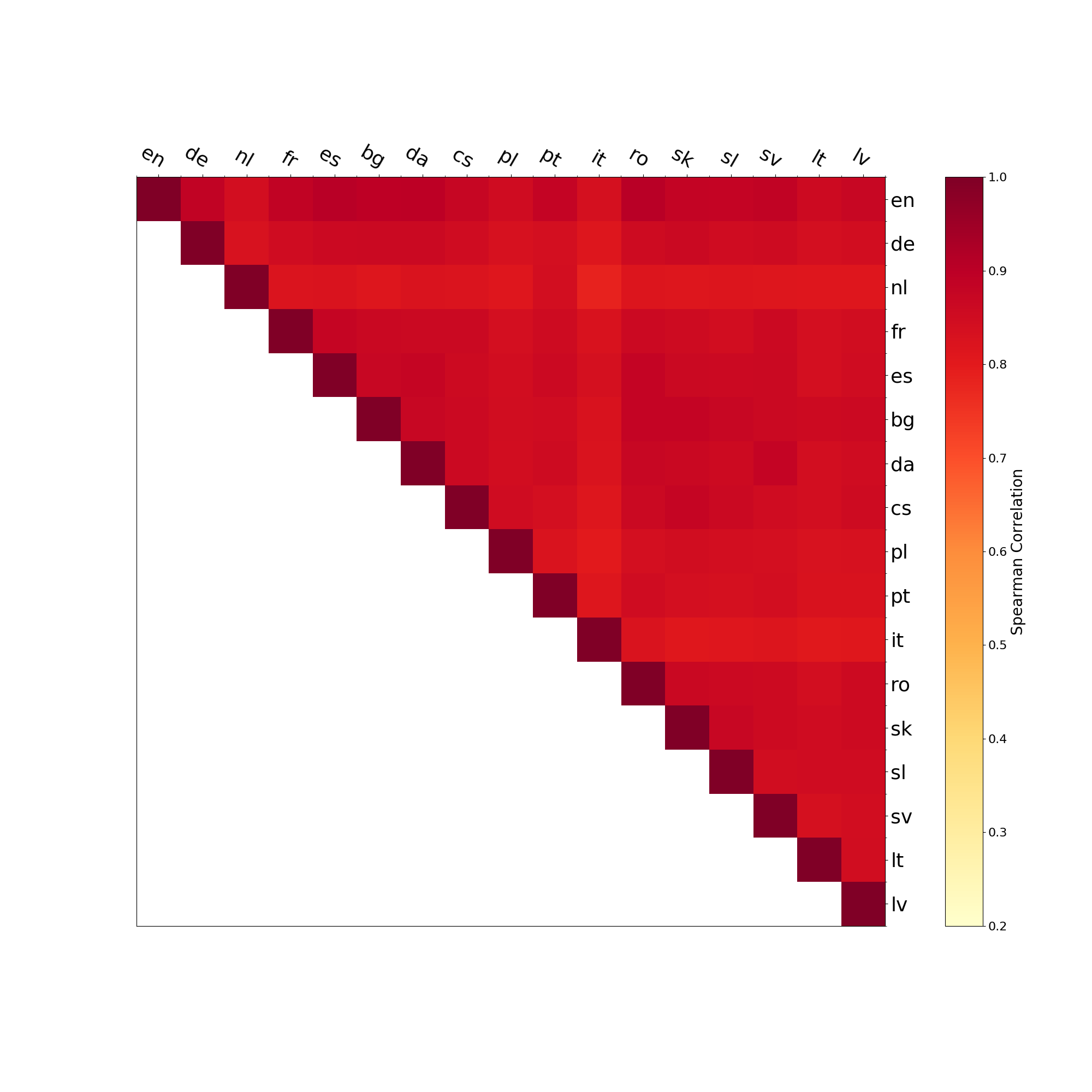}
\caption{Full representational similarity analysis for the \textit{mid} Europarl sentences.}
\label{fig:RSAEuroparl}
\end{figure}
\begin{figure}
\centering
  \includegraphics[width=\textwidth]{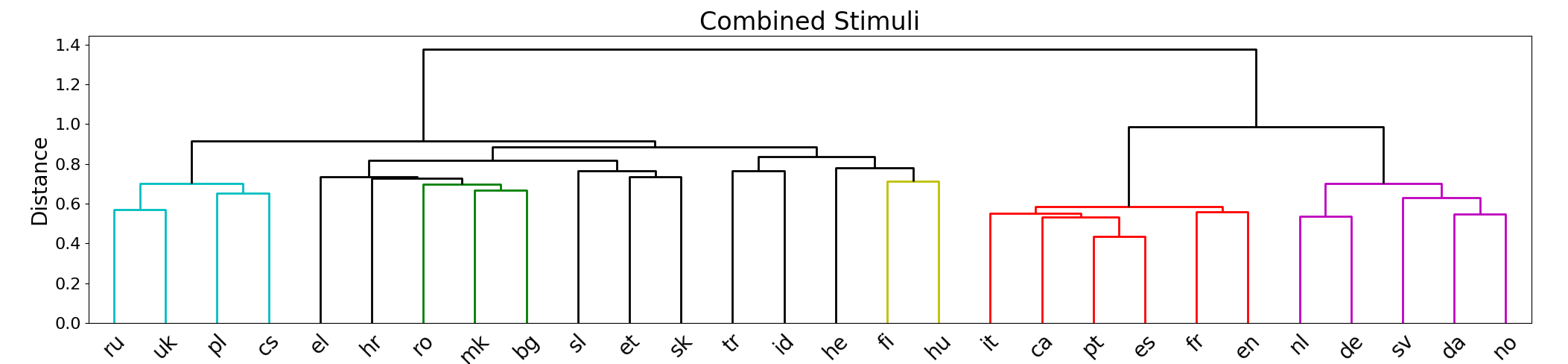}
\caption{The clustering tree that emerges when the \textsc{Swadesh} and \textsc{Pereira} stimuli are combined. }
\label{fig:dendrogram_combined}
\end{figure}
\end{document}